\documentclass{article}
\usepackage{PRIMEarxiv}
\usepackage[utf8]{inputenc} % allow utf-8 input
\usepackage[T1]{fontenc}    % use 8-bit T1 fonts
\usepackage{hyperref}       % hyperlinks
\usepackage{url}            % simple URL typesetting
\usepackage{booktabs}       % professional-quality tables
\usepackage{amsfonts}       % blackboard math symbols
\usepackage{nicefrac}       % compact symbols for 1/2, etc.
\usepackage{microtype}      % microtypography
\usepackage{lipsum}
\usepackage{graphicx}       % graphics
\usepackage{amsmath,amssymb}

\title{Affect-Conditioned Image Generation}

\author{
  Francisco Ibarrola\\
  The University of Sydney,\\
  NSW, Australia.\\
  \scalebox{.8}[1.0]{\texttt{francisco.ibarrola@sydney.edu.au}}\\
  \And
  Rohan Lulham\\
  The University of Sydney,\\
  NSW, Australia.\\
  \scalebox{.8}[1.0]{\texttt{rohan.lulham@sydney.edu.au}}\\
  \And
  Kazjon Grace \\
  The University of Sydney,\\
  NSW, Australia.\\
  \scalebox{.8}[1.0]{\texttt{kazjon.grace@sydney.edu.au}} \\
}

\begin{document}

\maketitle

\begin{figure*}[!h]
\centering
  \begin{minipage}{0.25\textwidth}
    \centering
    \it{happy (high Valence)}
    \includegraphics[width=\textwidth]{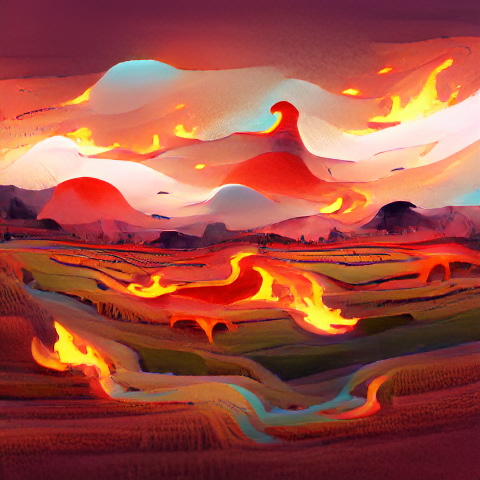}
    \end{minipage}
    \hspace{0.07\textwidth}
    \begin{minipage}{0.25\textwidth}
    \centering
    \it{excited (high Arousal)}
    \includegraphics[width=\textwidth]{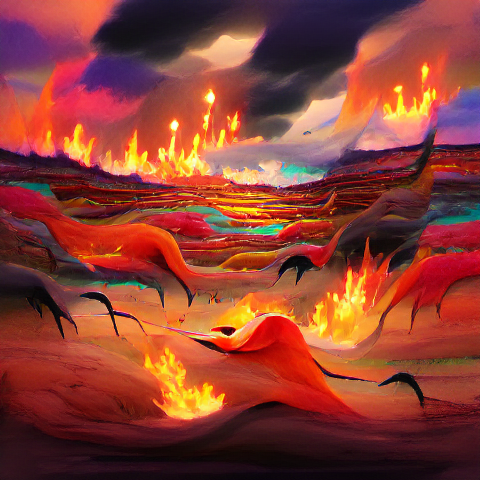}       
    \end{minipage}
    \hspace{0.07\textwidth}
    \begin{minipage}{0.25\textwidth}
    \centering
    \it{in control (high Dominance)}
    \includegraphics[width=\textwidth]{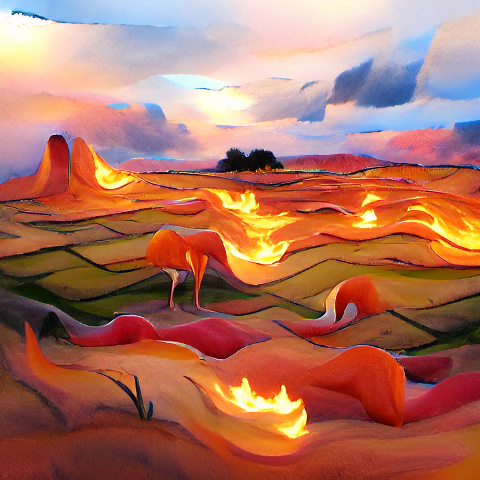}
    \end{minipage}
    
    \smallskip

    \begin{minipage}{0.25\textwidth}
    \centering
    \it{unhappy (low Valence)}
    \includegraphics[width=\textwidth]{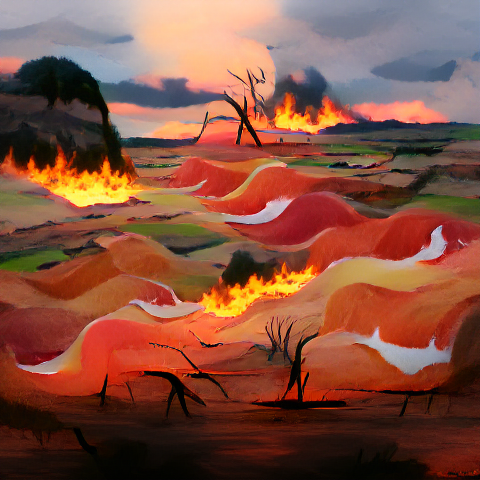}   
    \end{minipage}
    \hspace{0.07\textwidth}
    \begin{minipage}{0.25\textwidth}
    \centering
    \it{calm (low Arousal)}
    \includegraphics[width=\textwidth]{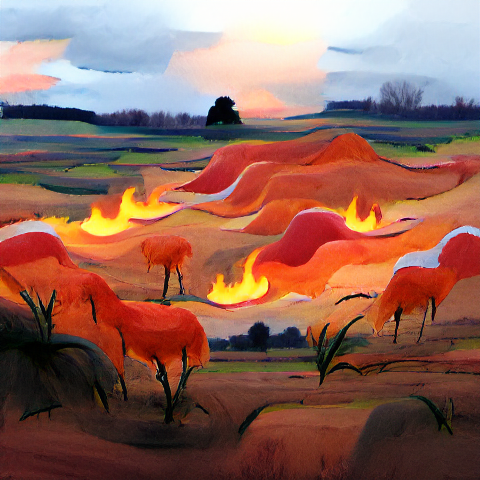}        
    \end{minipage}
    \hspace{0.07\textwidth}
    \begin{minipage}{0.25\textwidth}
    \centering
    \scalebox{.95}[1.0]{\it{not in control (low Dominance)}}
    \includegraphics[width=\textwidth]{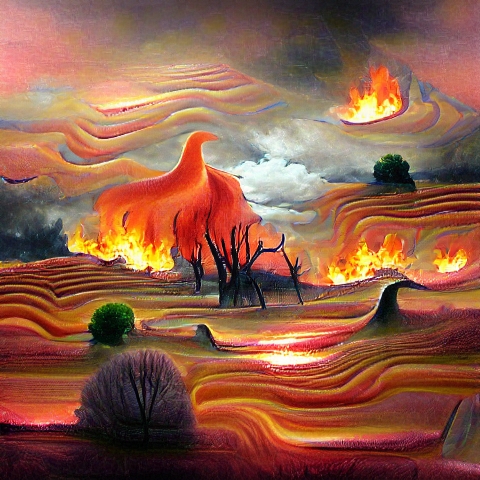}
    \end{minipage}
    % \captionsetup{width=\textwidth}
    \label{fig:vqgan_affect_flaming_landscape}
    \caption{Example images of a ``flaming landscape'' generated using VQGAN+CLIP with our affect conditioning, maximising (top row) and minimising (bottom row) the three dimensions we utilise: valence (left), arousal (middle), and dominance (right).}
\end{figure*}

\begin{abstract}
In creativity support and computational co-creativity contexts, the task of discovering appropriate prompts for use with text-to-image generative models remains difficult. In many cases the creator wishes to evoke a certain impression with the image, but the task of conferring that succinctly in a text prompt poses a challenge: affective language is nuanced, complex, and model-specific. In this work we introduce a method for generating images conditioned on desired affect, quantified using a psychometrically validated three-component approach, that can be combined with conditioning on text descriptions. We first train a neural network for estimating the affect content of text and images from semantic embeddings, and then demonstrate how this can be used to exert control over a variety of generative models. We show examples of how affect modifies the outputs, provide quantitative and qualitative analysis of its capabilities, and discuss possible extensions and use cases.
\end{abstract}

%%%%%%%%%%%%%%%%%%%%%%%%%%%%%%%%%%%%%%%%%
%%%%%%%%%%%%%%%%%%%%%%%%%%%%%%%%%%%%%%%%%
\section{Introduction}
%%%%%%%%%%%%%%%%%%%%%%%%%%%%%%%%%%%%%%%%%
%%%%%%%%%%%%%%%%%%%%%%%%%%%%%%%%%%%%%%%%%

Recent advances in Generative AI have made it possible to produce high quality images based solely on a textual description \cite{crowson2022vqganclip,frans2021clipdraw,saharia2022imagen}. These techniques are largely based on learning a latent space that encodes the semantic meanings of words or phrases, which can then be used to condition the generative process. Images typically contain much more information than short text prompts, and hence generative models need to learn the distribution of the residual information in order to ``fill in the gaps''. As a result, controlling this behaviour through iterative prompt engineering has  become a major component of real-world workflows using these technologies to produce images \cite{liu2022design}. 

These difficulties are often encountered within an artistic or design setting, where users may seek to produce an image that evokes a certain impression in its viewers. Choosing the right words to do that is a significant problem, and one that extends beyond the obvious challenge of coming up with the right words to convey a feeling, as complex or highly specific prompts can suffer from adverse effects like visual concept leakage. For example, ``a man stuck in a traffic jam is eating'' will, in DALLE-2 \cite{ramesh2022hierarchical}, produce images of a man in a car eating \textit{jam straight from the jar} \cite{rassin2022dalle}. Such surprising images are unlikely to match a user's intention (even if in this case the scene is borderline plausible). The complexity of the landscape of semantic latent spaces and the sparseness of the conditioning stimuli can lead to undesirable outputs, particularly when prompts are grammatically complex or involve words with multiple meanings.

Mismatches between intent and output are particularly relevant when human-AI co-creation is involved \cite{ruiz2022dreambooth,ibarrola2022cicada}, due to the constructive nature of the creative process itself. Psychological studies of creativity suggest that the tasks to which creative thinking is applied are under-specified \cite{paton2011briefing}, meaning that creativity is as much a matter of problem-framing as of problem-solving \cite{maher1996modeling}. A user occupied with wrestling an obstinate model to capture what they want right now, we contend, will be less able to refine those intents and figure out what they actually want. In creativity support applications such as art (where it may be desirable that an image evoke a particular mood or emotion \cite{cook2019framing}), or design (where diverse solutions must be explored \cite{grace2017personalised}), this significantly damages the utility of existing generative models. An ethnographic study of an online community that sprung up around a popular text-to-image generator supports this: the intent-output mismatch was found to be a significant practical challenge, so much so that the time-intensive process of collaborative prompt engineering became a major focus for the community \cite{oppenlaender2022creativity}.

In this paper we propose an additional conditioning stream, intended to be a simple and abstract way to diminish the need for highly-specific iteratively-tuned prompts, and in doing so reduce this mismatch between user and model semantics. We propose to condition images directly on affect: the emotive component of experiencing a stimulus. Images with very similar semantic content ("a dog in a forest") can evoke very different affect based on details not present in the prompt: perhaps the dog is snarling, perhaps it's an injured puppy, or perhaps it's excitedly retrieving a thrown stick. All of those details could be added to the prompt, but only if the user already knew exactly what they wanted, which is often not the case in creativity-support \cite{cascini2022perspectives}.

We develop a method for conditioning image generation in a variety of models, where affect is quantified using a psychometrically validated approach \cite{osgood1975cross}. This research established that cross-culturally, peoples' feelings about a wide range of stimuli (including words, phrases, scenarios, images, and people) varied on the same three affective dimensions; Valance (AKA evaluation, or how ``nice'' or ``happy'' something feels), Arousal (AKA activity, or how ``exciting'' or ``lively'' something feels) and Dominance (AKA power, or how ``in control'' something makes one feel). In Affect Control Theory, large data sets are used to model social interaction, experience and uncertainty with impressive fidelity \cite{heise2007expressive,warriner2013words,hoey2015bayesian}.  Computational models of these dimensions have been used in sentiment analysis tasks (typically just the valence and arousal dimensions, \cite{wang2016dimensional}), but not -- to the best of our knowledge -- to condition generative models. 

We show that these three dimensions of affect (which we call VAD scores, for Valence, Arousal \& Dominance) can be predicted from CLIP \cite{radford2021clip} and BERT \cite{devlin2018bert} encodings, and that this can be used to guide generative models. We demonstrate this in both generation-by-optimization models (CLIPDraw \cite{frans2021clipdraw} and VQGAN+CLIP \cite{crowson2022vqganclip}) and in conditional generation (Stable Diffusion \cite{rombach2022stablediffusion}). We show through a user study that our affect-conditioning allows for generating images with desired affect content without degrading the quality of images (i.e. fit-to-prompt). We end with a qualitative analysis of the relationship between semantics and affect-guided generation, as well as a discussion of the potential applications of this technique. Code to reproduce our experiments is available at \href{https://github.com/fibarrola/affect_conditioned_generation}{https://github.com/fibarrola/affect\_conditioned\_ generation}

%%%%%%%%%%%%%%%%%%%%%%%%%%%%%%%%%%%%%%%%%
%%%%%%%%%%%%%%%%%%%%%%%%%%%%%%%%%%%%%%%%%
\section{Affect Prediction}\label{sec:aff_pred}
%%%%%%%%%%%%%%%%%%%%%%%%%%%%%%%%%%%%%%%%%
%%%%%%%%%%%%%%%%%%%%%%%%%%%%%%%%%%%%%%%%%

In order to build a model capable of generating images conditioned on affect scores, we begin by training a neural network to estimate such scores. It is timely to note that our available data for this task consist of a list of 13k+ words, whose affect scores were derived through surveys \cite{warriner2013words}, and a dataset of 1.1k+ images \cite{lang1999international}, also with their respective VAD scores.

We begin by recalling that CLIP \cite{radford2021clip} consists of two neural networks $f$ and $g$ that map images and text (respectively) to a common latent space $\mathcal{Z}$, where the cosine similarity between two points attest to how semantically close they are. This means that given a (square) image $x\in[0,1]^{3\times M\times M}$ and a caption $t$, their cosine similarity in $\mathcal{Z}=\mathbb{R}^{512}$, $\langle g(t), f(x) \rangle$ should be large if and only if the caption corresponds to the image (and vice versa). By transitivity, this also implies that semantically similar images (or text prompts) should be close in $\mathcal{Z}$.

Since images and text are indistinguishable in $\mathcal{Z}$, and so are the scores in $\mathbb{R}^3$, the problem can be reduced to training a neural network $A:\mathbb{R}^{512}\rightarrow\mathbb{R}^3$. Given that there is no evident structural correlation between the coordinates of $\mathcal{Z}$, we can simply use a Multi-Layer Perceptron (MLP), whose details and training parameters are specified in Sections \ref{sec:net_arq} and \ref{sec:net_training}.

%%%%%%%%%%%%%%%%%%%%%%%%%%%%%%%%%%%%%%%%%
\subsection{Network Architecture}\label{sec:net_arq}
%%%%%%%%%%%%%%%%%%%%%%%%%%%%%%%%%%%%%%%%%

Our affect prediction model had an input dimension of 512 (to match the CLIP latent space), followed by two rectified linear layers of dimension 64 and 32, followed by an output layer of dimension 3 (to match the VAD model of affect). All layers included bias nodes.

%%%%%%%%%%%%%%%%%%%%%%%%%%%%%%%%%%%%%%%%%
\subsection{Data and Training}\label{sec:net_training}
%%%%%%%%%%%%%%%%%%%%%%%%%%%%%%%%%%%%%%%%%

Training was performed using a dataset of 13913 words and another of 1194 images, both annotated with affect scores $v\in [0, 8.6]^3$. We computed their CLIP embeddings using the corresponding networks, and combined them into a single dataset of 15107 vectors in $\mathbb{R}^{512}$. The dataset was split $70\%/30\%$ into training and testing sets, and training was performed for 1500 epochs, backpropagating using Adam optimization and with $20\%$ dropout between layers.

Previous to training, we scaled the CLIP embeddings as well as the affect scores to $[0,1]$. We shall hereafter refer to the affect scores on the $[0,1]$ interval for ease of understanding.

%%%%%%%%%%%%%%%%%%%%%%%%%%%%%%%%%%%%%%%%%
%%%%%%%%%%%%%%%%%%%%%%%%%%%%%%%%%%%%%%%%%
\section{Affect-Conditioned Generation}
%%%%%%%%%%%%%%%%%%%%%%%%%%%%%%%%%%%%%%%%%
%%%%%%%%%%%%%%%%%%%%%%%%%%%%%%%%%%%%%%%%%

Having developed a neural network $A$ that can predict affect scores based on CLIP latents, we can use it to condition image generation models to produce outputs with a desired affect score $v_0\in\mathbb{R}^3$.

Let us consider the space of square images of $M$ by $M$ pixels $\mathcal{X}\doteq [0,1]^{3\times M \times M},$ and an image generator $h:\mathcal{Y}\rightarrow\mathcal{X}$ that produces an image $x\in \mathcal{X}$ based on a set of parameters $y\in\mathcal{Y}$.

%%%%%%%%%%%%%%%%%%%%%%%%%%%%%%%%%%%%%%%%%
\subsection{Generation-by-optimization with affect}\label{sec:aff_ctrl_gen_by_opt}
%%%%%%%%%%%%%%%%%%%%%%%%%%%%%%%%%%%%%%%%%

Some generative models, like VQGAN+CLIP \cite{crowson2022vqganclip} and CLIPDraw \cite{frans2021clipdraw}, work by guiding $h$ by optimizing the set of parameters $y$ for the produced image to match a given text prompt $t$ in terms of CLIP embeddings. This means, finding $y$ minimizing a semantic loss
\begin{align*}
    L_s(g(t), f\circ h(y)),
\end{align*}
 where as before $g$ and $f$ are CLIP's text and image encoders, respectively.

In VQGAN+CLIP, $\mathcal{Y}$ is VQGAN's \cite{esser2021vqgan} latent space and $L_s$ is the spherical distance between the CLIP embeddings of the prompt and generated image. In CLIPDraw, $\mathcal{Y}$ is the set of possible B\'ezier curves parameters, and $L_s$ is the cosine similarity between the CLIP embeddings of the prompt and generated image.

Given that our affect-predicting neural network $A$ is a differentiable function, we can guide these kind of generation-by-optimization methods by defining a new cost function
\begin{align}\label{eqn:gen_by_opt_loss}
    L_{go}(y; t, v_0) \doteq & L_s(g(t), f\circ h(y)) +\lambda\|A\circ f\circ h(y) - v_0\|^2,
\end{align}
where $\lambda>0$ is a regularization parameter that determines the weight we assign to the output matching the affect vector against the semantic meaning of the prompt.

%%%%%%%%%%%%%%%%%%%%%%%%%%%%%%%%%%%%%%%%%
\subsection{Conditional generative models with affect}
\label{sec:aff_ctrl_st_diff}
%%%%%%%%%%%%%%%%%%%%%%%%%%%%%%%%%%%%%%%%%

In other generative models, such as Stable Diffusion \cite{rombach2022stablediffusion}, the generator $h$ takes an embedding of the text prompt $z=b(t)$ as conditioning input, and the result depends solely on this and the initial conditions or noise $\eta$.

In the particular case of Stable Diffusion, the embedding space $\mathcal{Z}$ consists of 77 channels of fine-tuned BERT embeddings. We know that close points in this space correspond to prompts with similar semantic meanings, and so a way to steer an output to have a desired affect score $v_0$ would be to find points close to $z(t)\in\mathbb{R}^{77\times 768}$ that have a more suitable score. Namely, minimizing with respect to $z$
\begin{align} \label{eqn:st_diff_loss}
L_{sd}(z; t, v_0) \doteq \sum_c \|b_c(t)-z_c\|^2 +\lambda\|A_c(z_c)-v_0\|^2,
\end{align}
where $\lambda>0$ is a weighting parameter and $A_c$ is a MLP trained (as before) to predict affect scores from the $c$-th channel of the embedding.

The network structure used for this was the same as the one described in Section \ref{sec:net_arq}, except the first layer input is of size 768 (to match a single BERT channel). Of course, this entails training 77 similar networks, but given the number of network parameters this is fairly fast. Unlike the CLIP space, BERT encodes only text, so we removed the affect-image pairs from our training data and just trained these 77 networks on (70\% of) the 13,913 affect-word pairs. The process is otherwise analogous to the previous one, and both code and our pre-trained models are available from the same repository.

%%%%%%%%%%%%%%%%%%%%%%%%%%%%%%%%%%%%%%%%%
%%%%%%%%%%%%%%%%%%%%%%%%%%%%%%%%%%%%%%%%%
\section{Results}
%%%%%%%%%%%%%%%%%%%%%%%%%%%%%%%%%%%%%%%%%
%%%%%%%%%%%%%%%%%%%%%%%%%%%%%%%%%%%%%%%%%

\begin{figure*}[h]
    \centering
    \setlength\tabcolsep{8pt}
    % @{\extracolsep{\fill}}
    % \scalebox{1}{``A smiling woman sitting on the beach''}
    \includegraphics[width=\textwidth]{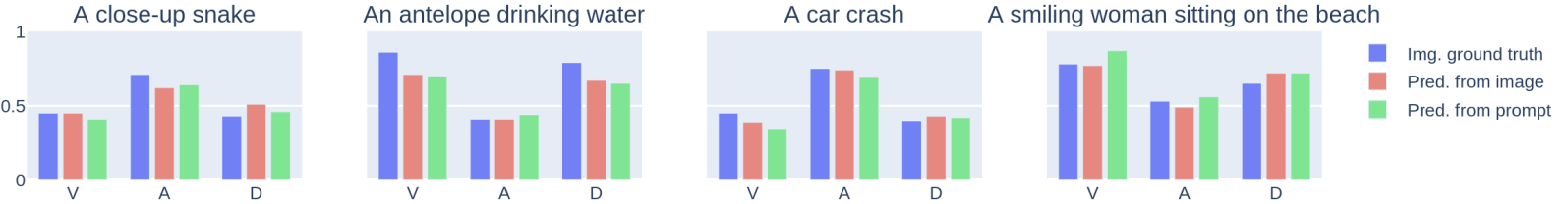}
    \caption{Affect values of images matching four prompts. In blue, the ground truth affect values for the images obtained in surveys from [Lang {\it et al.}, 1999]. In red, the affect scores estimated from the images using the model described in Section \ref{sec:aff_pred}, in green the affect scores estimated from the prompts by the same model. The licence for using these images prohibits their publication to preserve their value.}
    \label{tab:prompt_affect_scores}
\end{figure*}

Let us start by assessing the performance of our affect-predicting neural networks, trained from CLIP latents as well as from BERT embeddings. 

\begin{table}[ht]
    \centering
    \setlength\tabcolsep{40pt}
    \begin{tabular*}{\columnwidth}{ l | c | c }
       Type &   Mean Error (CLIP) & Mean Error (BERT)\\
        \hline
       Text    &   $0.064$     & $0.062 \pm 0.007$\\
       Image   &   $0.076$     & N/A \\      
    \end{tabular*}
    \caption{Estimation errors over the test data for our affect scores prediction networks, based on CLIP latents (images and text) or BERT embeddings (77 channels, text).}
    \label{tab:mlp_errors}
\end{table}

The obtained training errors are depicted in Table \ref{tab:mlp_errors}. These are computed over affect scores scaled to $[0, 1]$, so we can think of them as percentages. Given that the ground truth data is from a survey, the target affect scores for a given word (or image) are the means of the empirical distributions of user ratings. In that sense it is worth mentioning that $97.7\%$ of estimations using CLIP fall within a standard deviation of the survey means, as well as $98.4\%$ of BERT-based predictions.

Note that the models were trained on single words (and images in the case of the CLIP version), but given that we are basing predictions on latent spaces capable of encoding both words and phrases, it should also work for predicting the affect scores of text prompts. While we do not have a ground truth for evaluating (or training) on multi-word input, we can approximately assess the model's performance by manually generating prompts that describe images in our training set (for which we do have VAD scores) and then comparing the model's predicted affect for that phrase to the ground-truth affect available for the image. This is imprecise as there are many ways of describing any image, but broad agreement would suggest efficacy of phrase-level prediction. Figure \ref{tab:prompt_affect_scores} shows brief descriptions of four randomly selected images (the actual images are under a non-disclosure policy) along with their true affect scores, the scores estimated from the images using the model described in Section \ref{sec:aff_pred}, and the affect scores predicted using the image descriptions as prompts, using the same model. Note that these text prompts were written before any predictions were made, without iteration or cherry-picking. We would include the images themselves, but the licence for using this dataset prohibits their publication to preserve their value for psychology experiments. Note that across all four prompts there are very small differences between ground truth, prediction from the prompt, and prediction from the images.

% \begin{table}[h]
%     \centering
%     \setlength\tabcolsep{4pt}
%     % @{\extracolsep{\fill}}
%     % \scalebox{1}{``A smiling woman sitting on the beach''}
%     \begin{tabular*}\columnwidth{l|c|c|c}
%         prompt & V & A & D \\
%         \hline
%        ``A close-up snake''  & $0.41$ & $0.64$ & $0.46$ \\
%        ``An antelope drinking water'' &     $0.70$ & $0.44$ & $0.65$ \\
%        ``A car crash'' & $0.34$ & $0.69$ & $0.42$ \\
%        ``A smiling woman sitting on the beach'' & $0.87$ & $0.56$ & $0.72$ \\
%     \end{tabular*}
%     \caption{Affect scores of text prompts.}
%     \label{tab:prompt_affect_scores}
% \end{table}

%%%%%%%%%%%%%%%%%%%%%%%%%%%%%%%%%%%%%%%%%
\subsection{Generation-by-optimization with affect}
%%%%%%%%%%%%%%%%%%%%%%%%%%%%%%%%%%%%%%%%%

\begin{figure*}[h]
    \centering
    \includegraphics[width =\textwidth]{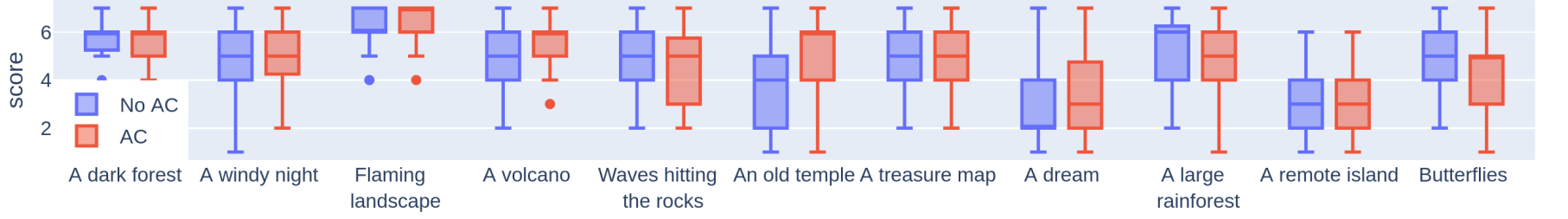}
    \vspace{-0.6cm}
    \caption{Quality (i.e. fit-to-prompt) ratings from a survey of 27 participants for images generated using generation-by-optimization methods, with (AC) or without (No AC) incorporating Affect Control. With no observable quality effects between the two we claim that our affect-conditioning approach does not impact subjective image quality.}
    \label{fig:survey_scores}
\end{figure*}

In order to test incorporating affect control into a generation-by-optimization context, we used the approach described in Section \ref{sec:aff_ctrl_gen_by_opt} with CLIPDraw and VQGAN+CLIP. That is, we chose the corresponding semantic loss $L_s$ in \ref{eqn:gen_by_opt_loss} and different prompts and target affect vectors $v$.

Figure 1 shows an example of images generated using VQGAN+CLIP with a fixed prompt (``flaming landscape'') and different target affect vectors. Target affect vectors were constructed to have a $1$ (or $0$) in the affect position we wanted to enhance (or decrease), and $0.5$ in the remaining components.

A first-order qualitative examination of the six images in Fig. 1 shows significant differences in tone and content. The high valence image has small, almost flighty wisps of flame and pleasantly contrasting sky colours.  The low valence image has dead trees, larger flames, and burnt-out areas.  The high arousal image has towering flames spitting pillars of black smoke over a rough landscape, while the low arousal image has a few small, contained flames and a blue sky. The high dominance image (remember that "dominance" here means the \textit{viewer} feels dominant or in control, not that the image dominates the viewer) is likewise calm with a blue sky, while the low-dominance image shows a large fire-front and a rough sky. All six images are of a ``flaming landscape'', but their affective content clearly varies.

In order to quantitatively support these observations, we conducted a survey that consisted of two tasks. First, the 27 participants were presented with a set of images (generated with or without affect control) and asked to rate how they matched a provided prompt, using a 7-pt Likert scale. This was to test whether our method entailed a loss of quality (i.e. fit-to-prompt). Secondly, they were given pairs of affect-conditioned images, one generated to have a high score of either valence, arousal, or dominance, and the other with a low score on the same component. Half the images were from VQGAN+CLIP \cite{crowson2022vqganclip}, the other half from CLIPDraw \cite{frans2021clipdraw}. The subjects were then asked to pick which image of the pair felt happier (if the conditioning component was Valence), or more calm (for Arousal) or made them feel more in control (for Dominance).

The results of the first (image quality) task for each prompt are depicted in Figure \ref{fig:survey_scores}. No consistent trend is observed across the prompts, suggesting that affect conditioning does not reduce subjective ratings of fit-to-prompt.

\begin{figure}
    \centering
    \includegraphics[width =0.6\columnwidth]{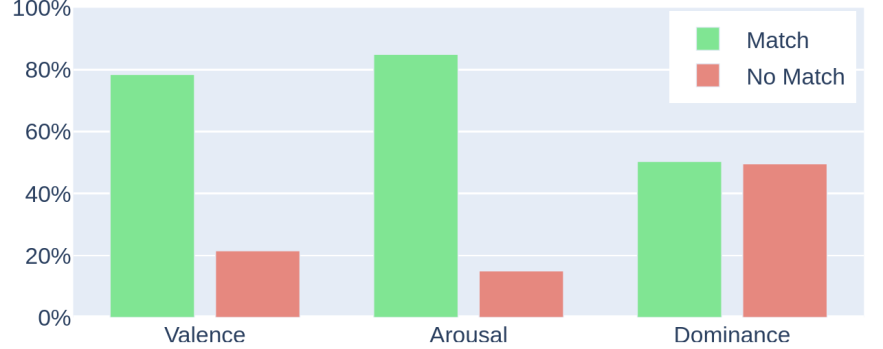}
    \caption{User choices regarding which images better reflect the corresponding affect state.}
    \label{fig:survey_preferences}
\end{figure}

Participant choices for perceived affect are summarized in Figure \ref{fig:survey_preferences}. Respondents could clearly distinguish affect-conditioned images for Valence and Arousal, but not for Dominance. This suggests that our affect-conditioning model aligns with viewer expectations across a variety of prompts and two different generators for the first two affect dimensions, but not for the third. Subjectively, participants reported difficulty understanding what it meant to feel ``in control'' or ``not in control'' of an image, potentially suggesting confounds in that part of the study. It's also important to note that the images generated by VQGAN+CLIP and CLIPDraw are abstract and dream-like, which may have been less compatible with feelings of dominance than with feelings of happiness/unhappiness or excitement/calm.

%%%%%%%%%%%%%%%%%%%%%%%%%%%%%%%%%%%%%%%%%
\subsection{Affect Conditioned Stable Diffusion}
%%%%%%%%%%%%%%%%%%%%%%%%%%%%%%%%%%%%%%%%%

Stable Diffusion produces images of higher quality than the synthesis-based methods, so we perform a  more in-depth qualitative analysis on results generated with it. We generated several paired samples of images conditioned to have low or high values of each affect component. The images are ``paired'' in the sense that the diffusion process was initiated from the same noise tensors, meaning that the only differences we should observe are a consequence of the affect conditioning. We present our analysis of several pairs here, noting that the nature of the subject matter makes that analysis tend unavoidably towards the poetic. Further examples can be found in the supplementary material.

\begin{figure*}[!h]
\centering
    \begin{minipage}[t]{0.02\textwidth}
    \rotatebox{90}{\hspace{1cm}High Valence}
    \end{minipage}
    \includegraphics[width=0.23\textwidth]{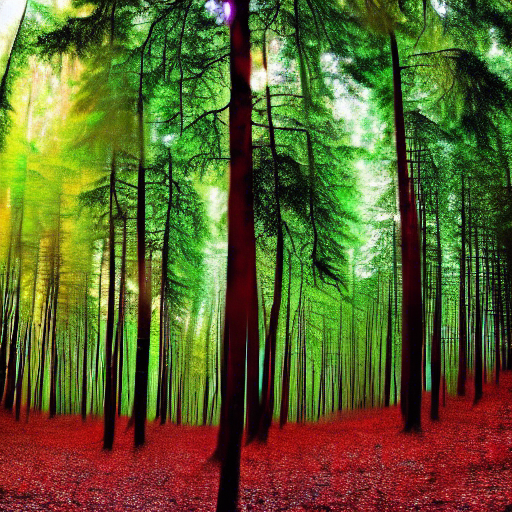}
    \hspace{0.008\textwidth}
    \includegraphics[width=0.23\textwidth]{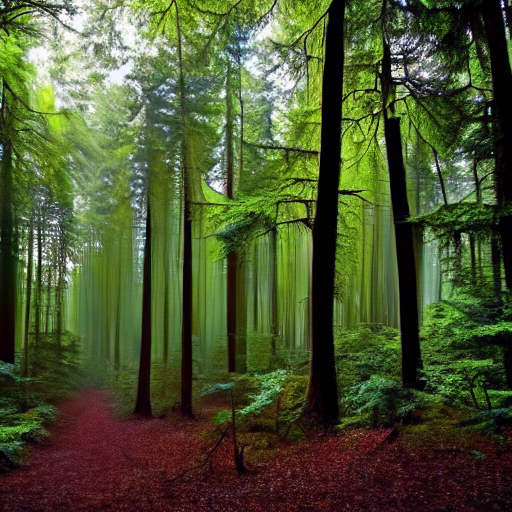}
    \hspace{0.008\textwidth}
    \includegraphics[width=0.23\textwidth]{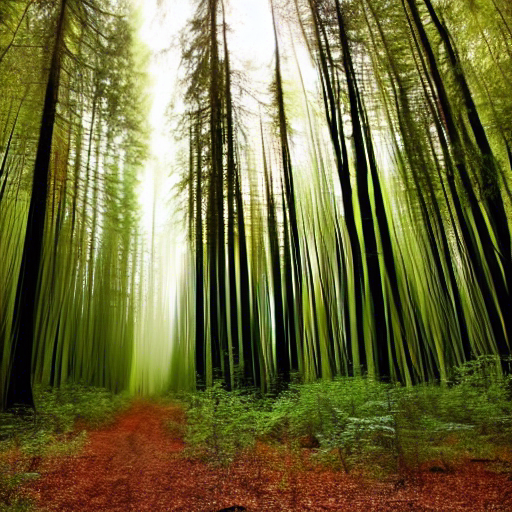} 
    \hspace{0.008\textwidth}
    \includegraphics[width=0.23\textwidth]{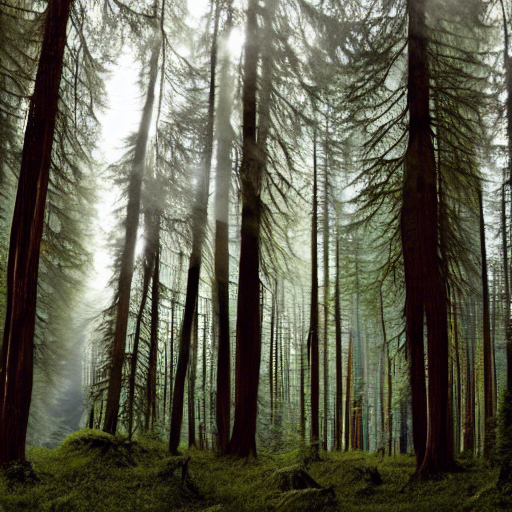}

    \bigskip
    
    \begin{minipage}[t]{0.02\textwidth}
    \rotatebox{90}{\hspace{1.1cm}Low Valence}
    \end{minipage} 
    \includegraphics[width=0.23\textwidth]{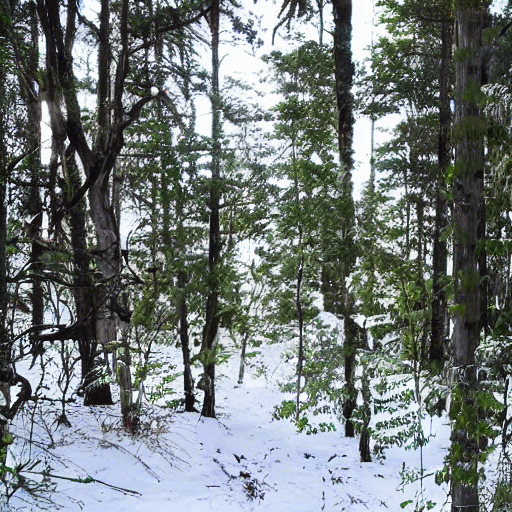}
    \hspace{0.008\textwidth}
    \includegraphics[width=0.23\textwidth]{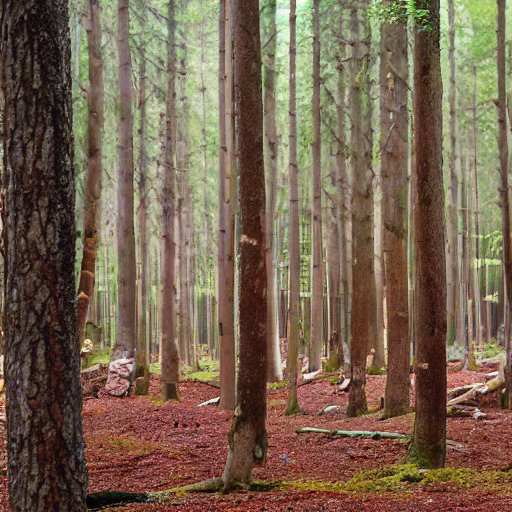}
    \hspace{0.008\textwidth}
    \includegraphics[width=0.23\textwidth]{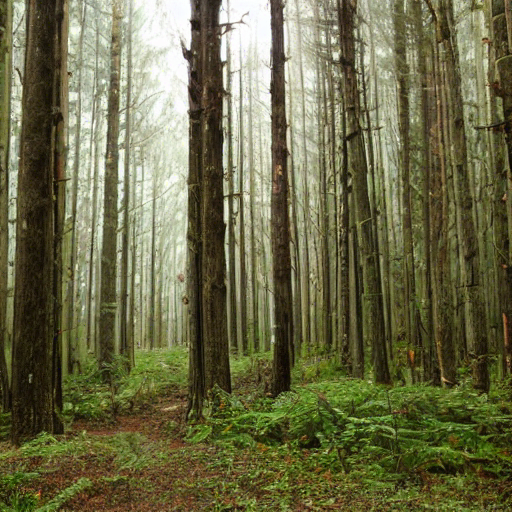}
    \hspace{0.008\textwidth}
    \includegraphics[width=0.23\textwidth]{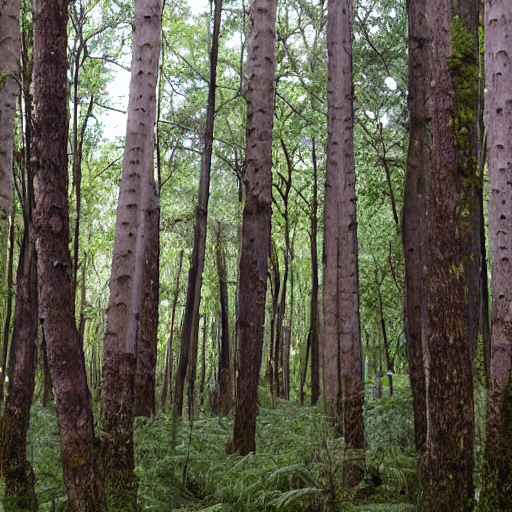}
    \caption{Images of ``A forest'' generated using affect-conditioned Stable Diffusion with either High or Low Valence.\smallskip}
    \label{fig:st_diff_eval}
\end{figure*}

The influence of Valence on the prompt ``A forest'' can be seen in Figure \ref{fig:st_diff_eval}. The high valence forests appear to have more sunlight, and thicker, greener foliage, which are known to be a human preference \cite{kuper2015preference}. One low-valence image, by contrast, is snow-covered. The viewpoint of the images tends to point upwards, emphasising the height of the trees in a way that we interpret as evoking majesty or reverence.  By contrast the low-valence images have a viewpoint that focuses on the depth of the forest, perhaps evoking a feeling of being lost or endangered.

\begin{figure*}[!h]
\centering
    \begin{minipage}[t]{0.02\textwidth}
    \rotatebox{90}{\hspace{1cm}High Arousal}
    \end{minipage}
    \includegraphics[width=0.23\textwidth]{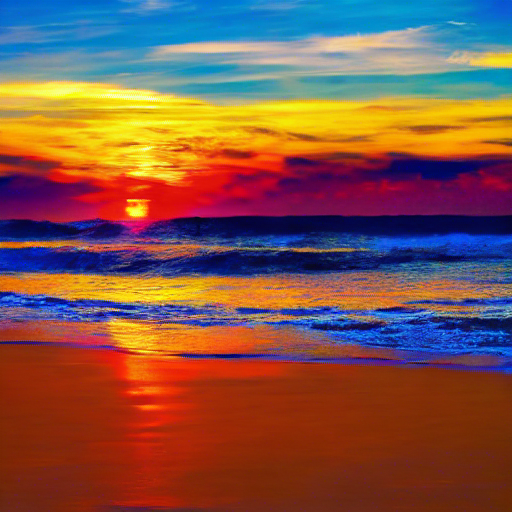}
    \hspace{0.008\textwidth}
    \includegraphics[width=0.23\textwidth]{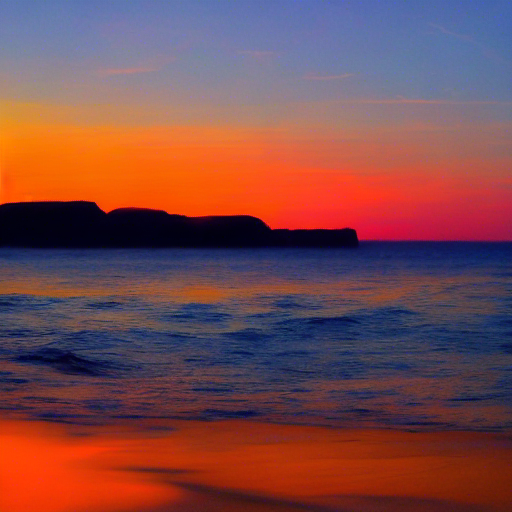}
    \hspace{0.008\textwidth}
    \includegraphics[width=0.23\textwidth]{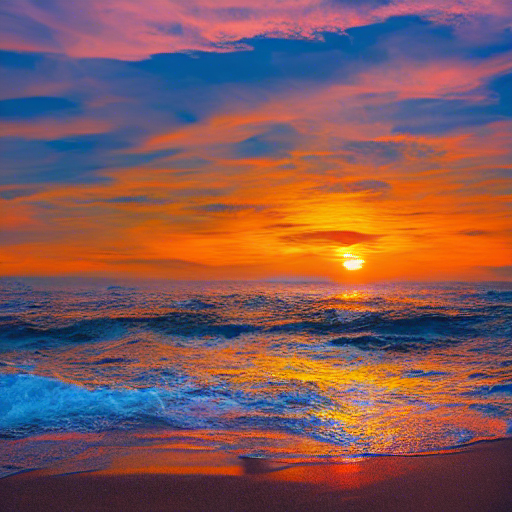}
    \hspace{0.008\textwidth}
    \includegraphics[width=0.23\textwidth]{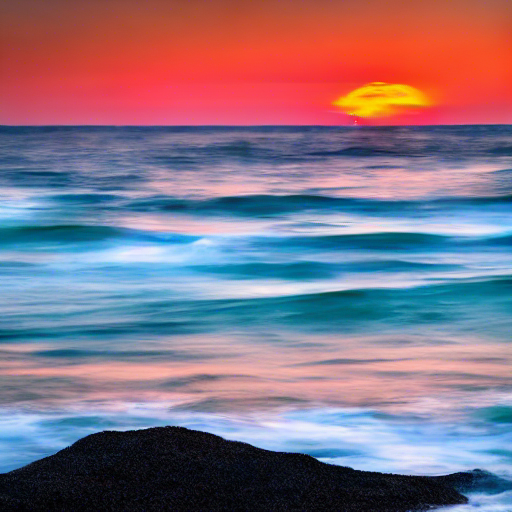} 

    \bigskip

    \begin{minipage}[t]{0.02\textwidth}
    \rotatebox{90}{\hspace{1cm}Low Arousal}
    \end{minipage}    
    \includegraphics[width=0.23\textwidth]{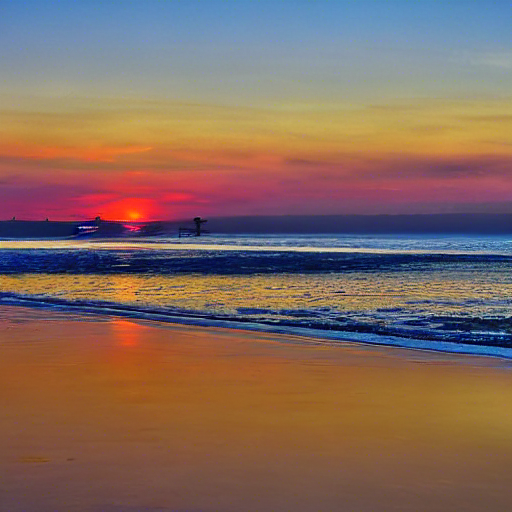}
    \hspace{0.008\textwidth}
    \includegraphics[width=0.23\textwidth]{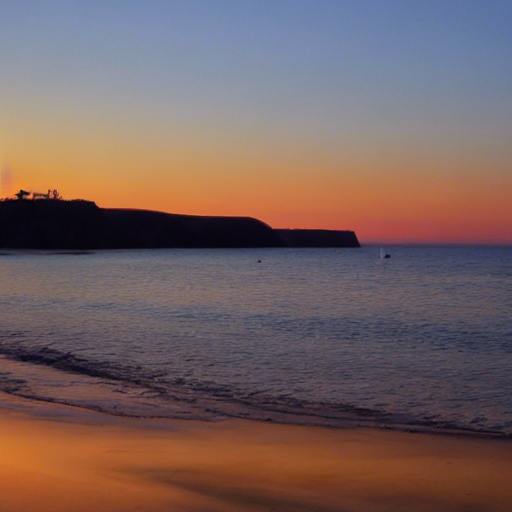}
    \hspace{0.008\textwidth}
    \includegraphics[width=0.23\textwidth]{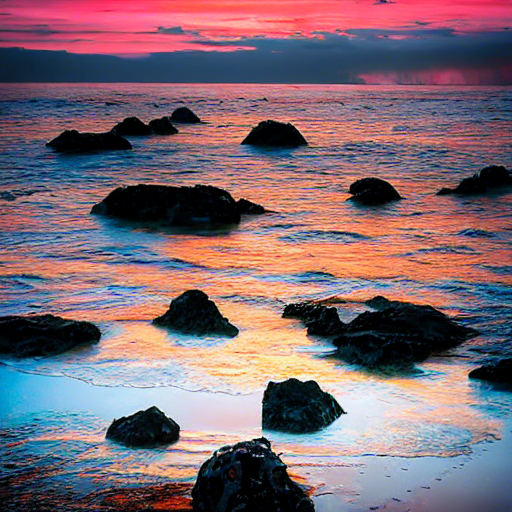}
    \hspace{0.008\textwidth}
    \includegraphics[width=0.23\textwidth]{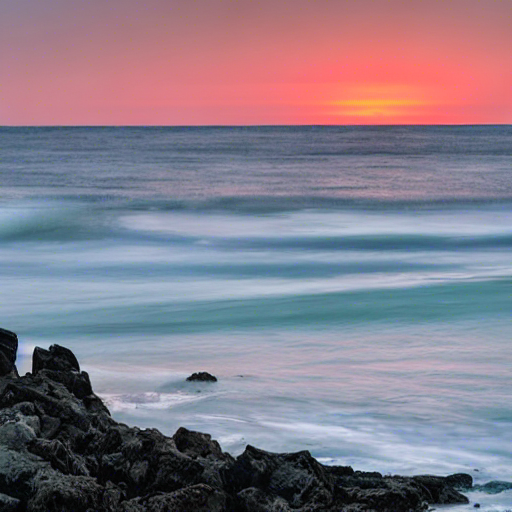} 
    \caption{Images of ``The sea at sunrise'' generated using affect-conditioned Stable Diffusion with either High or Low Arousal.\smallskip}
    \label{fig:st_diff_potency}
\end{figure*}

How Arousal affects the prompt ``the sea at sunrise'' (Figure \ref{fig:st_diff_potency}) is more direct. The lower Arousal images depict a flat sea with small, even waves, consistent with the notion that ``still waters create feelings of calmness and tranquility.'' \cite{sakici2015assessing}. In addition to the rougher seas, high arousal also appears to correlate with a stronger and more vibrant dawn light, with greater contrast accentuating the clouds.

\begin{figure*}[!h]
\centering
    \begin{minipage}[t]{0.02\textwidth}
    \rotatebox{90}{\hspace{0.15cm}High (viewer) Dominance}
    \end{minipage}
    \includegraphics[width=0.23\textwidth]{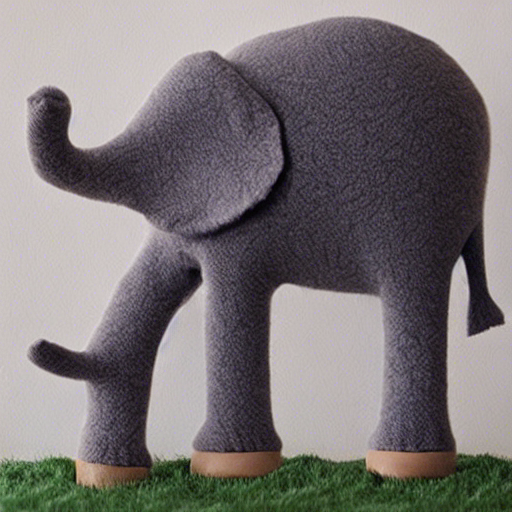}
    \hspace{0.008\textwidth}
    \includegraphics[width=0.23\textwidth]{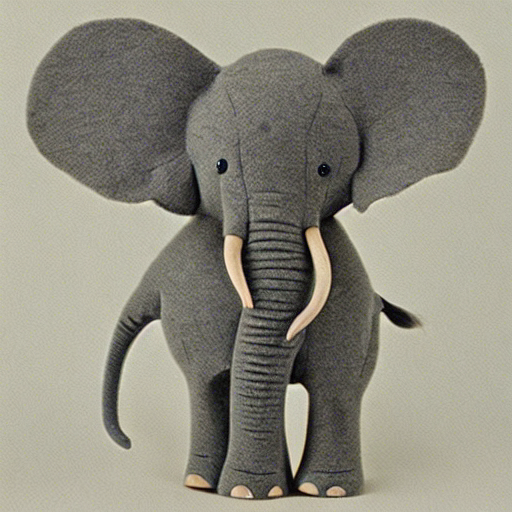}
    \hspace{0.008\textwidth}
    \includegraphics[width=0.23\textwidth]{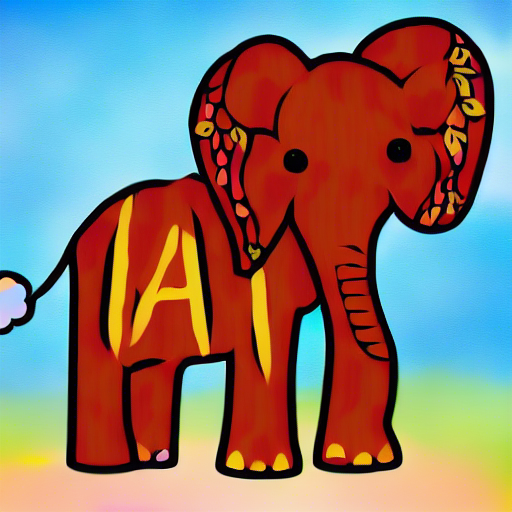} 
    \hspace{0.008\textwidth}
    \includegraphics[width=0.23\textwidth]{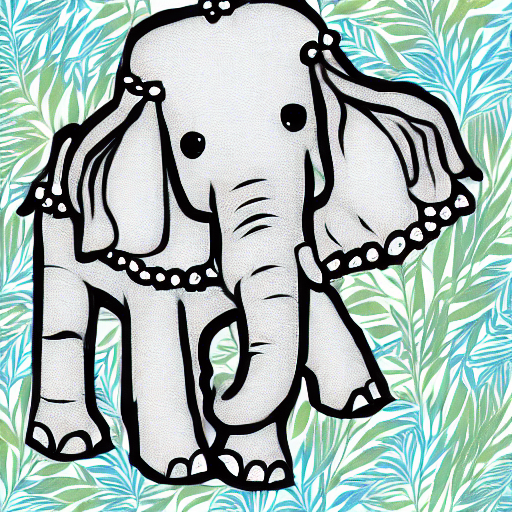}
    
    \bigskip
    
    \begin{minipage}[t]{0.02\textwidth}
    \rotatebox{90}{\hspace{0.2cm}Low (viewer) Dominance}
    \end{minipage}
    \includegraphics[width=0.23\textwidth]{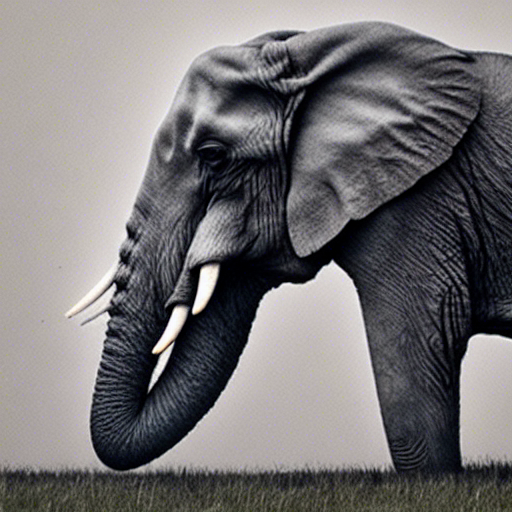}
    \hspace{0.008\textwidth}
    \includegraphics[width=0.23\textwidth]{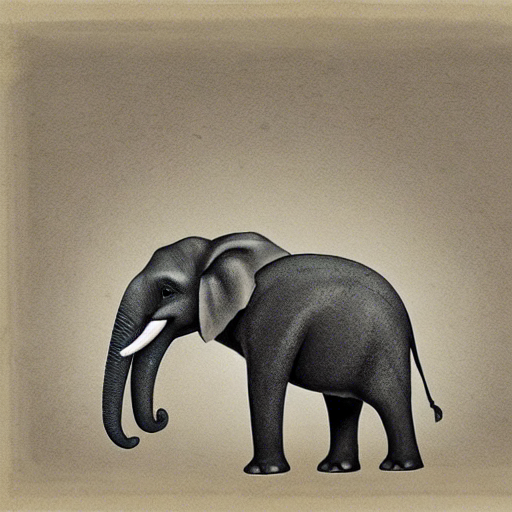}
    \hspace{0.008\textwidth}
    \includegraphics[width=0.23\textwidth]{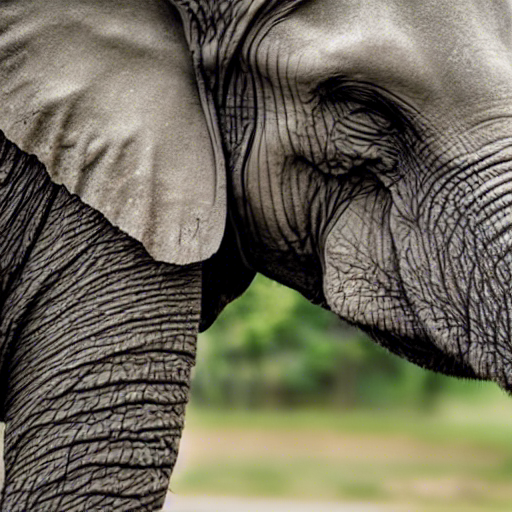}
    \hspace{0.008\textwidth}
    \includegraphics[width=0.23\textwidth]{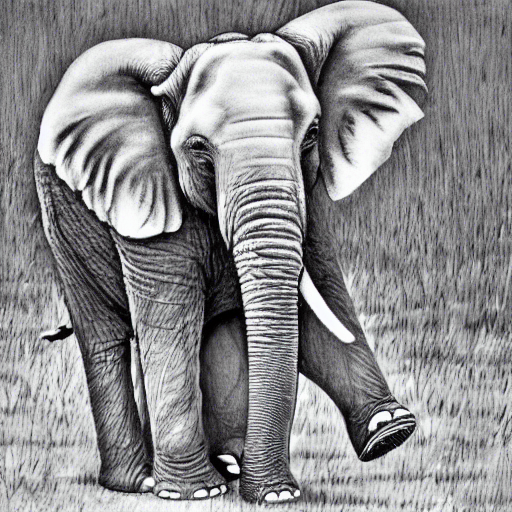} 

    \caption{Images generated using affect-conditioned Stable Diffusion using the prompt ``An elephant'' with either High or Low Dominance.\smallskip}
    \label{fig:st_diff_activity}
\end{figure*}

Unlike what we have observed with the methods that generate abstract images, the effect of dominance appears clearly in the images of "An elephant" in Figure \ref{fig:st_diff_activity}. Feeling in control watching something as imposing as an elephant seems unlikely, but our method produced images of children's toys or cartoons when high Dominance was enforced. This qualitative difference shows the capacity of affect-conditioning to push generators into diverse and a-priori unlikely regions of the subspaces of images associated with a particular prompt.

This suggests that affect-controlled image generation may be useful for inspiring new ways in which we can confer a desired affect to a design or artwork. Such modifications could then be integrated into the prompt in a creative application, furthering that iterative process of discovery.  To demonstrate this, we modified the prompts for the above images using characteristics we observed in the affect-conditioned results. Table \ref{tab:modified_prompt_affect_scores} shows our model's predicted affect scores for each of those prompt modifications, showing consistently larger or smaller affect scores in the corresponding components as those modifiers are added or removed.  This shows the potential of affect conditioning both as a tool not just for producing images that provoke a particular affect directly, but also for exploring the space of appropriate prompts.

\begin{table}[h]
    \centering
    \setlength\tabcolsep{40pt}
    \begin{tabular*}\columnwidth{l|c|c|c}
        Prompt & V & A & D \\
        \hline
    A forest in summer & \textbf{0.802} & 0.408 & 0.647 \\
    A forest & \textbf{0.746} & 0.407 & 0.646 \\
    A forest in winter & \textbf{0.695} & 0.411 & 0.572 \\
    \hline
    A wavy sea at sunrise & 0.680 & \textbf{0.365} & 0.602 \\
    The sea at sunrise & 0.770 & \textbf{0.356} & 0.641 \\
    A calm sea at sunrise & 0.768 & \textbf{0.331} & 0.655 \\
    \hline
    A toy elephant & 0.610 & 0.376 & \textbf{0.591} \\
    An elephant & 0.610 & 0.370 & \textbf{0.577} \\
    A real elephant & 0.554 & 0.400 & \textbf{0.545} \\
    \end{tabular*}
    \caption{Affect scores of text prompts, computed from CLIP latents.}
    \label{tab:modified_prompt_affect_scores}
\end{table}

%%%%%%%%%%%%%%%%%%%%%%%%%%%%%%%%%%%%%%%%%
%%%%%%%%%%%%%%%%%%%%%%%%%%%%%%%%%%%%%%%%%
\section{Discussion and Conclusions}
%%%%%%%%%%%%%%%%%%%%%%%%%%%%%%%%%%%%%%%%%
%%%%%%%%%%%%%%%%%%%%%%%%%%%%%%%%%%%%%%%%%

\begin{figure*}[!h]
\centering
    \begin{minipage}[t]{0.02\textwidth}
    \rotatebox{90}{\hspace{1cm}High Valence}
    \end{minipage}
    \includegraphics[width=0.23\textwidth]{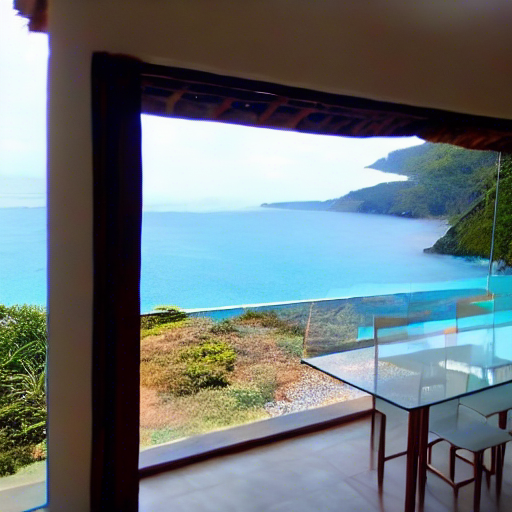}
    \hspace{0.008\textwidth}
    \includegraphics[width=0.23\textwidth]{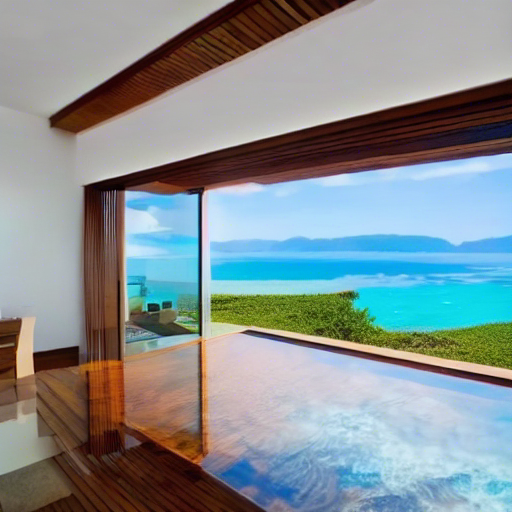}
    \hspace{0.008\textwidth}
    \includegraphics[width=0.23\textwidth]{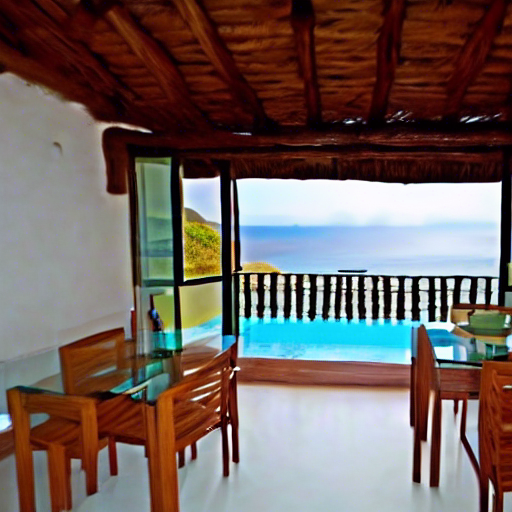} 
    \hspace{0.008\textwidth}
    \includegraphics[width=0.23\textwidth]{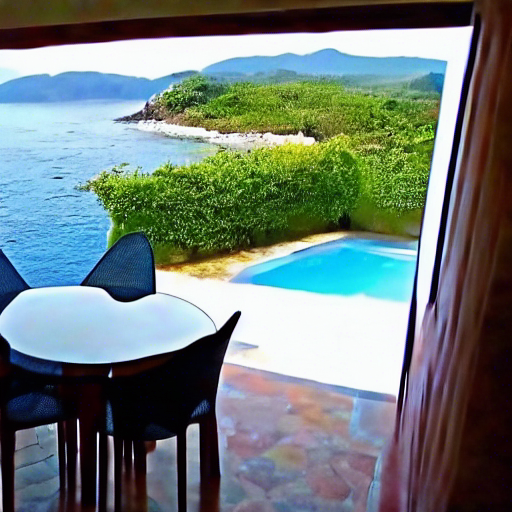}
    
    \bigskip

    \begin{minipage}[t]{0.02\textwidth}
    \rotatebox{90}{\hspace{1.1cm}Low Valence}
    \end{minipage} 
    \includegraphics[width=0.23\textwidth]{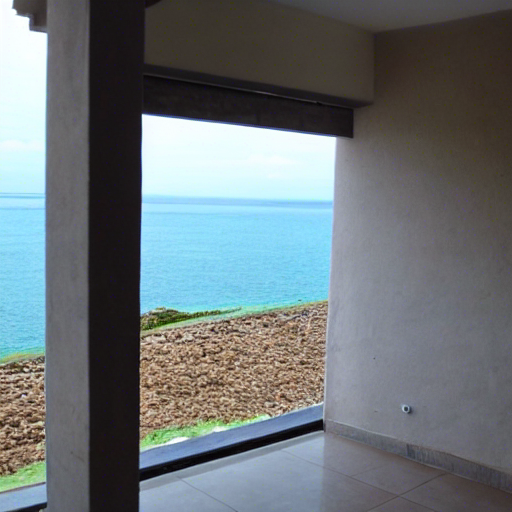}
    \hspace{0.008\textwidth}
    \includegraphics[width=0.23\textwidth]{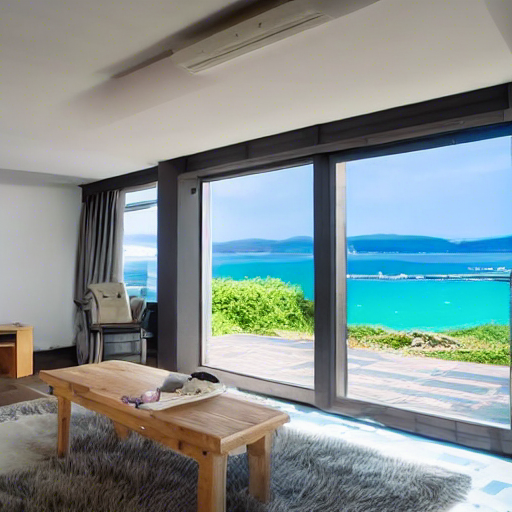}
    \hspace{0.008\textwidth}
    \includegraphics[width=0.23\textwidth]{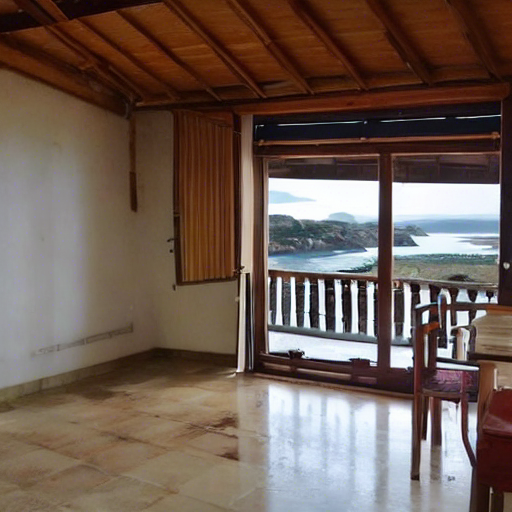}
    \hspace{0.008\textwidth}
    \includegraphics[width=0.23\textwidth]{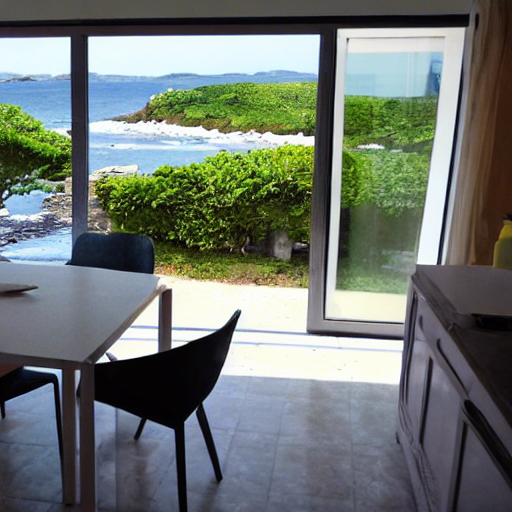} 
    \caption{Images of ``A house overlooking the sea'' generated using affect-conditioned Stable Diffusion with either High or Low Valence.\smallskip}
    \label{fig:st_diff_valence_house}
\end{figure*}

The success of our affect-conditioned generative models shows that CLIP semantic latents effectively represent affective information, in that it can be predicted from them (at least in the Valence and Arousal dimensions) accurately. This provides an interesting insight on the continuity of this latent space and its mapping to affect, since 90\% of our our training data consisted of single words, and that the remaining images constitute a relatively small number of data points compared to the number of learned parameters. 

%Upon looking at the results, we can firstly state that it is possible to estimate the affect scores of images from their CLIP semantic latents. This is somewhat surprising, in that the number of training latent vectors that come from images is small, relative to the size of the Network. This means that the continuity of the latent space is allowing the proposed Network to effectively learn mostly from the latents of single word embeddings, that constitute more than 90\% of the data.

Our survey suggests that affect conditioning can imbue Valence and Arousal scores into CLIPDraw and VQGAN+CLIP generated images. Our survey did not support a similar conclusion for the third component of affect, Dominance (although we note that previous research has characterised dominance as the "least strong" of the three \cite{lang1999international}). A qualitative analysis of our high- and low-dominance conditioned images (see Figure 1) does show some potential effects, but they were not human-discriminable in our study.  We hypothesize that this has to do with the inherent difficulty of conferring dominance in abstract images like those produced by VQGAN+CLIP and CLIPDraw. This conjecture is supported by affect-conditioned images from the less-abstract Stable Diffusion: while we did not conduct a second survey using these, the fact that it produced toys when told to draw an elephant that the viewer felt dominant over (see Figure \ref{fig:st_diff_activity} shows its power in at least some scenarios. 

%The results obtained with affect-conditioned Stable Diffusion show that the proposed approach applied to modifying text latents produces interesting results, that could even serve the purpose of prompt exploration. The tricks of reinterpreting an elephant as a toy elephant to increase the feel of control, or to focus on the tree tops rather than the trunks to increase pleasantness may not have been obvious before actually looking at the images.

We contend that affect conditioning is a valuable direction for future generative models, particularly in human-AI co-creative contexts.  Prompt engineering is hard and time consuming, because many creators do not know exactly what they are looking for \cite{paton2011briefing}, and even if they did the model may not interpret their words as they intend \cite{oppenlaender2022creativity}. We propose, however, that many artists and designers have a sense of the "vibe" they desire in their finished product, and affect conditioning gives a way for them to directly target that. To give a concrete example, users may want to increase control but may have not thought of tricks like drawing the elephant as a toy, or may want to increase pleasantness but not thought of focusing on the treetops rather than the horizon. 

Figure \ref{fig:st_diff_valence_house} shows another example of how affect-conditioning could guide design. It contains several versions of ``A house overlooking the sea'' with either high or low Valence. From these images a designer could infer certain desirable characteristics: having big windows, wood and glass furnishings, and a pool. While these may seem obvious to any architect or real-estate agent, we would emphasise that (since the affect prediction networks here were trained on BERT embeddings only), this model has never ``seen'' an image of ``a house overlooking the sea'' associated with any affect value. Those desirable visual features were inferred using only the continuity of the BERT latent space, with no images ever provided as reference.

While these results are promising, there is still much to do in this direction. It should be possible to re-train a diffusion-based model to directly incorporate affect, using the prediction network proposed in this work as ground truth, leading to faster (and potentially more accurate) real-time results. User studies of new generative creativity support tools that incorporate affect conditioning will also be needed to support our hypotheses about our approach's efficacy in that context.

\section*{Acknowledgments}

We would like to acknowledge the support of the Australian Research Council (Grant \#DP200101059) in completing this work.


\bibliographystyle{unsrt}

\end{document}